\documentclass{article}

\usepackage{microtype}
\usepackage{graphicx}
\usepackage{float}
\usepackage{booktabs} 
\usepackage{algorithmic}
\usepackage{algorithm}
\usepackage{amsmath}
\usepackage{amsthm}
\usepackage{amssymb}
\usepackage{multirow}
\usepackage{MnSymbol}
\usepackage{adjustbox}
\usepackage{enumitem}
\usepackage{balance}
\usepackage{subfig} 
\usepackage[english]{babel}
\usepackage{natbib}

\usepackage{hyperref}

\newcommand{\reffig}[1]{Figure~\ref{#1}}
\newcommand{\reftab}[1]{Table~\ref{#1}}
\newcommand{\refsec}[1]{Section~\ref{#1}}
\newcommand{\refeq}[1]{(\ref{#1})}
\newcommand{\refag}[1]{Algorithm \ref{#1}}
\newcommand{\refprop}[1]{Proposition~\ref{#1}}
\newcommand{\Real}{\mathbb{R}}
\newcommand{\reffigsub}[1]{\ref{#1}}

\newtheorem{propt}{Proposition}

\usepackage[accepted]{icml2021}

\icmltitlerunning{A Simple and Efficient Stochastic Rounding Method for Training Neural Networks in Low Precision}
\begin{document}

\twocolumn[
\icmltitle{A Simple and Efficient Stochastic Rounding Method for \\ Training Neural Networks in Low Precision}



\icmlsetsymbol{equal}{*}

\begin{icmlauthorlist}
	\icmlauthor{Lu Xia}{tue}
	\icmlauthor{Martijn Anthonissen}{tue}
	\icmlauthor{Michiel Hochstenbach}{tue}
	\icmlauthor{Barry Koren}{tue}
\end{icmlauthorlist}

\icmlaffiliation{tue}{Department of Mathematics and Computer Science, Eindhoven University of Technology, Eindhoven, The Netherlands}

\icmlcorrespondingauthor{Lu Xia}{l.xia1@tue.nl}

\icmlkeywords{Low-Precision Computation, Fixed-Point Arithmetic, Neural Networks, Binary Classification, Stochastic Rounding}

\vskip 0.3in
]



\printAffiliationsAndNotice{}  

\begin{abstract}
Conventional stochastic rounding (CSR) is widely employed in the training of neural networks (NNs), showing promising training results even in low-precision computations. We introduce an improved stochastic rounding method, that is simple and efficient.
The proposed method succeeds in training NNs with 16-bit fixed-point numbers and provides faster convergence and higher classification accuracy than both CSR and deterministic rounding-to-the-nearest method. 
\end{abstract}

\section{Introduction}
\label{sec:introduction}
In many computations, rounding is an unavoidable step, due to the finite-precision arithmetic supported in hardware. When simulating arithmetic in software, rounding is required when finite-precision arithmetic is to be used for reasons of computational efficiency. Many rounding schemes have been proposed and studied for different applications \cite{8766229}, such as round down, round up, round to the nearest and stochastic rounding. These rounding modes normally have different round-off errors. When a sequence of computations is implemented, round-off errors may be accumulated and magnified. In real-world problems, the magnification of round-off errors may cause divergence of numerical methods or mismatch between different arithmetic. When very high accuracy is required, data representation and arithmetic operations of higher precision are employed, for which computing times may be long. 

To reduce computing times and hardware complexity, and to increase the throughput of arithmetic operations, low-precision numerical formats are becoming increasingly popular, especially in the area of machine learning. An unbiased stochastic rounding scheme was applied by \citet{gupta2015deep} to train neural networks (NNs) using low-precision fixed-point arithmetic. The experiments show that when the rounding-to-the-nearest (RN) method fails, the training results using 16-bit fixed-point representation with the stochastic rounding method are very similar to those computed in 32-bit floating-point (single) precision. Inspired by \citet{gupta2015deep}, this conventional stochastic rounding (CSR) is widely employed in training NNs in low-precision floating-point or fixed-point computation, see, e.g., \citet{na2017chip,ortiz2018low,wang2018training}. \citet{nagel2020up} introduce an adaptive rounding method for post-training quantization of NNs by analyzing the rounding problem for a pre-trained NN. Additionally, the implementation of CSR in hardware is also growing \cite{8259423,su2020towards,mikaitis2020stochastic}. It has been stated that the failure of RN in training NNs is mainly caused by the loss of gradient information during the parameter updating procedure, because the weight updates that are below the minimum rounding precision are rounded to zero \cite{hohfeld1992probabilistic,gupta2015deep}. The gradient information is captured partially when CSR is applied. Still, whether it is worth to decrease the rounding bias by sacrificing gradient information is to be discussed.

Binary classifiers are widely applied in medical image classification, where deep NNs (DNNs) are shown to efficiently compute the final classification labels with raw pixels of medical images \cite{li2014medical,pan2015brain,lai2018medical}. Still, the DNN models suffer from high computational cost due to the high resolution of the medical images. Binary classifiers are also popular for other applications, such as classification of images for 3D scanning \cite{vezilic2017binary} and gender classification on real-world face images \cite{shan2012learning}. Again, computational costs are high due to large image data size. 

In this paper, we introduce an improved stochastic rounding method to train low-precision NNs for binary classification problems, which we call random rounding (RR). It provides more gradient information with a limited rounding bias in training NNs. The experiments show that the NNs trained using limited precision with RR lead to faster convergence rate and higher classification accuracy than those trained using RN and CSR. Furthermore, a constant rounding probability is applied in RR, which may significantly simplify the computational complexity and hardware implementation. We study the performance of the proposed RR first in dot product operations and then to train NNs with different rounding precision. 

The remainder of the paper is organized as follows. Standard deterministic rounding methods, the CSR method and the proposed RR method, are presented in \refsec{sec: section2}. \refsec{sec:section4} outlines the main mathematical operations in training NNs. In \refsec{sec:numericalstudy}, different rounding methods are studied in computing dot products and training NNs in limited precision. Finally, conclusions are drawn in \refsec{sec:conclusion}.

\section{Rounding Schemes}\label{sec: section2}
\begin{figure*}[ht]
	\vskip 0.15in
	\begin{center}
		\subfloat[RN]{\label{fig:VarianceandBiasforminimizedvariance3}\includegraphics[width=0.7\columnwidth]{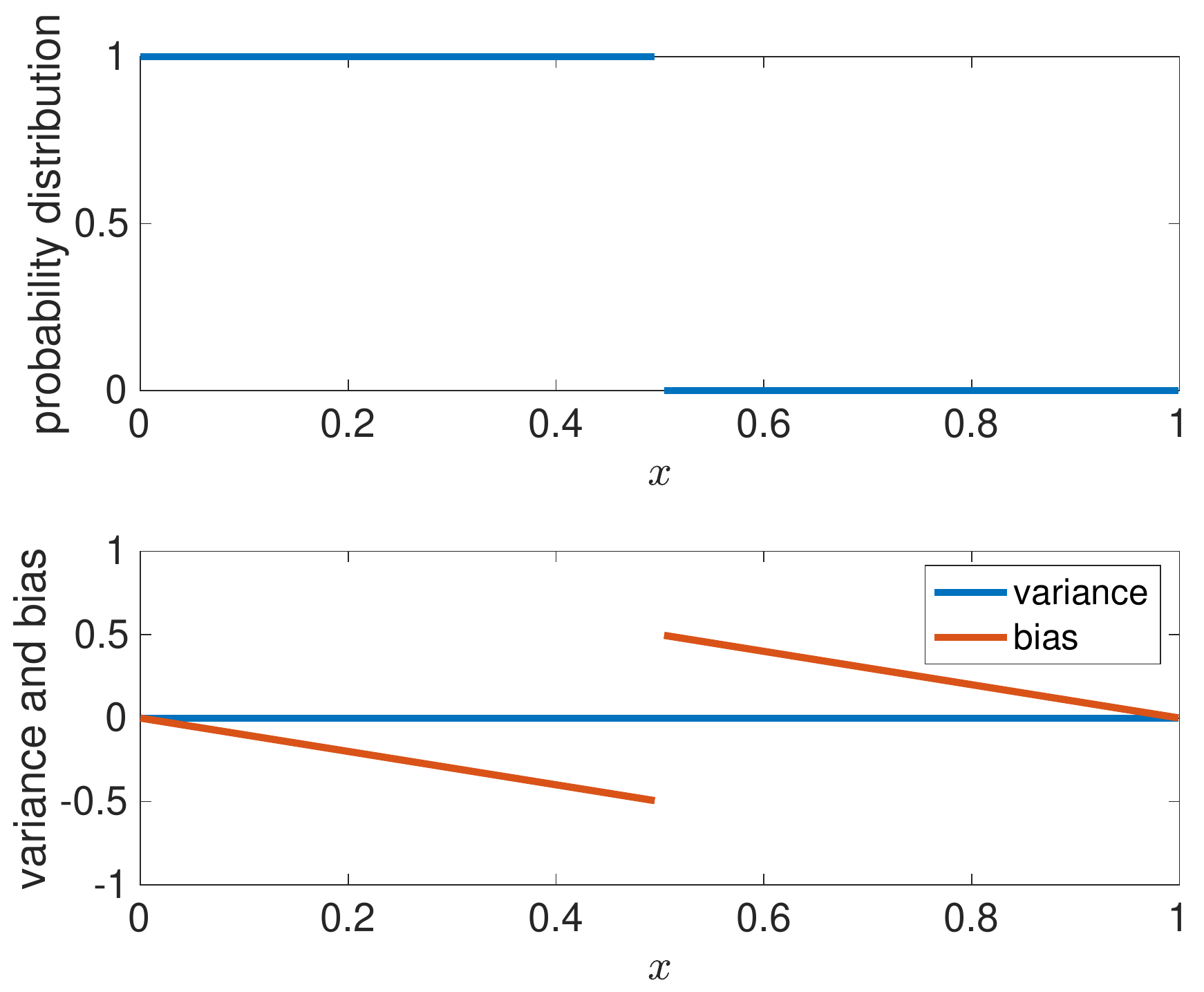}}
		\subfloat[CSR]{\label{fig:VarianceandBiasforStochasticrounding}\includegraphics[width=0.7\columnwidth]{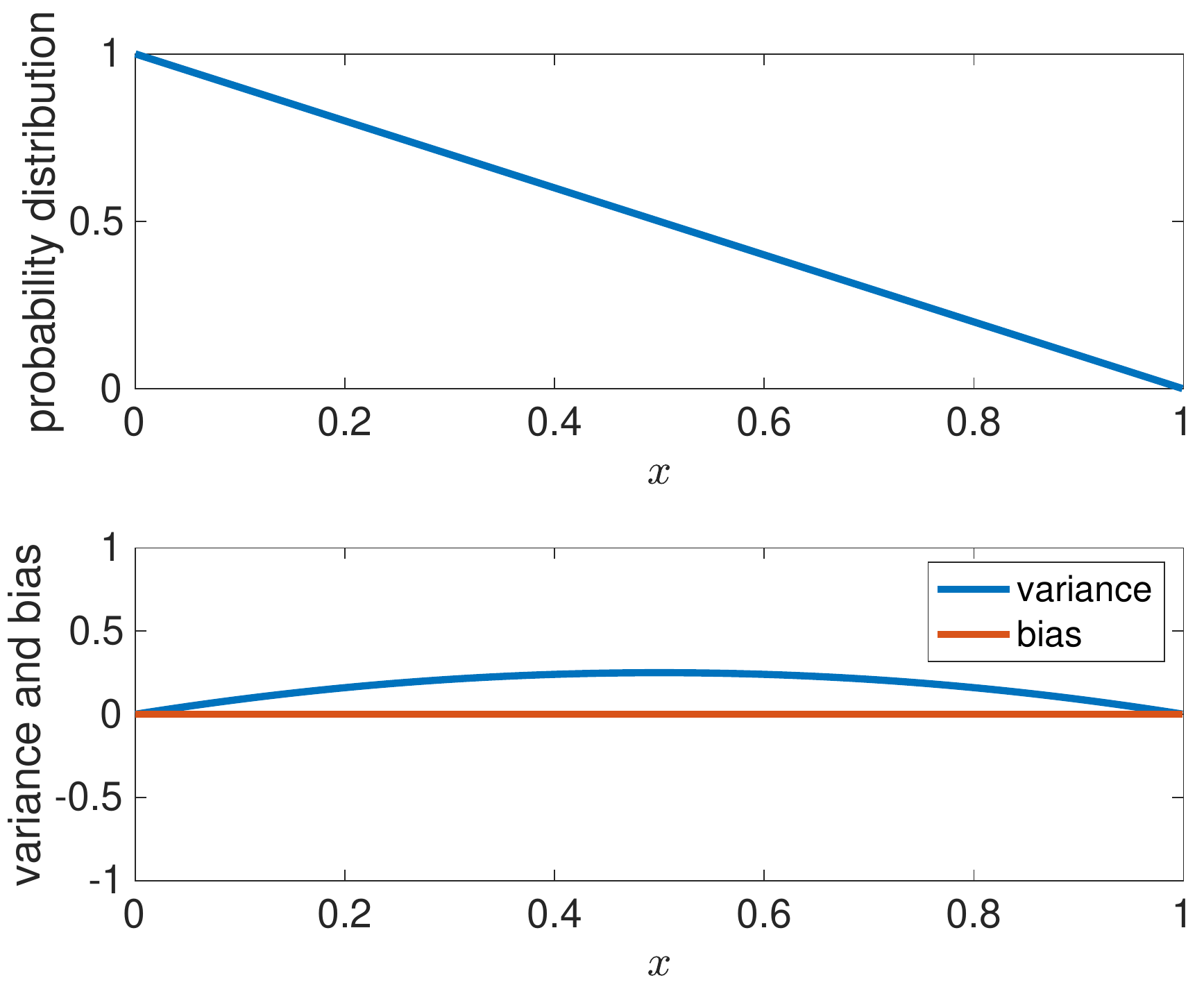}}
		\subfloat[RR]{\label{fig:SR2}\includegraphics[width=0.7\columnwidth]{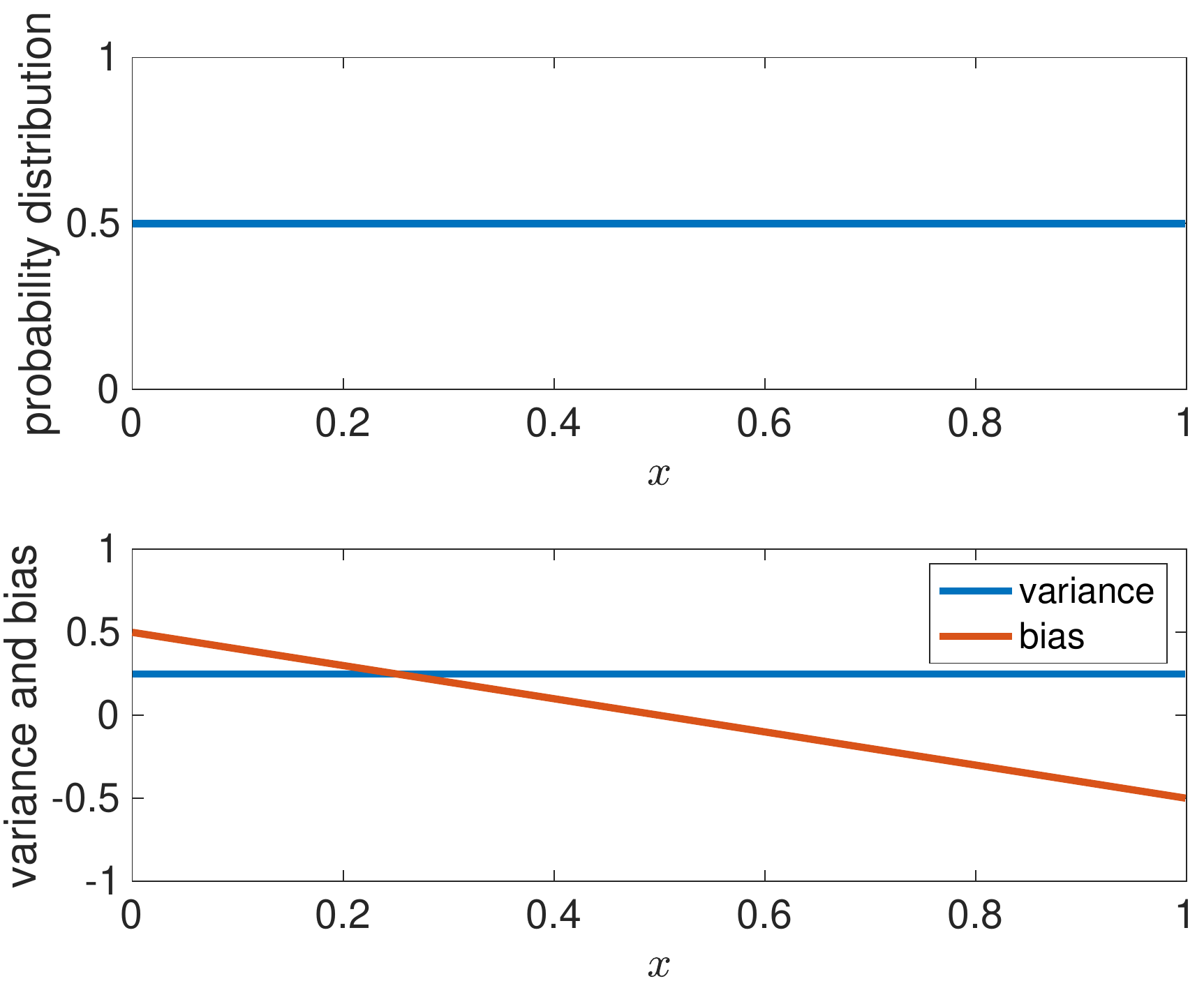}} 
		\caption{Probability distribution (top) and corresponding variance and bias (bottom) with respect to input variable $x\in [0,1)$ and $\delta=1$, for RN (a), CSR (b) and RR (c).}
		\label{fig:VarianceandBiasforminimizedvariance}
	\end{center}
	\vskip -0.12in
\end{figure*}
According to \citet{gupta2015deep}, training NNs using RN with a limited precision leads to a large degradation in classification accuracy from single-precision computation. It has been stated that the main reason of this failure is that RN fails to capture the gradient information during rounding, since it always rounds the numbers in $[-\frac{\delta}{2},\frac{\delta}{2}]$ to zero, where $\delta$ is the rounding precision \cite{hohfeld1992probabilistic,gupta2015deep}. In this section, we introduce an improved stochastic rounding method that has small probability of rounding to zero for numbers in $[-\frac{\delta}{2},\frac{\delta}{2}]$, while the rounding bias and variance are bounded.
\subsection{Introduction}
All the rounding schemes we study can be formulated as:
\begin{align}
	\mathrm{R}(x)= \begin{cases}
		\lfloor x \rfloor, \quad  \quad  &\text{with probability}~p(x),\\
		\lfloor x \rfloor+\delta,  \quad   &\text{with probability}~1-p(x),
	\end{cases}
	\label{eq:strounding}
\end{align}
with $p(x)\in[0,1]$, where $\mathrm{R}(x)$ is the rounded value of $x$ and $\lfloor x \rfloor$ indicates the largest representable floating-point or fixed-point number less than or equal to $x$. The variance of rounding scheme \refeq{eq:strounding} is 
\begin{equation}
	\mathit{V} = (\lfloor x \rfloor-\mu)^2p+(\lfloor x \rfloor+\delta-\mu)^2(1-p),
	\label{eq:variancestr1}
\end{equation}
where $\mu$ is the expected rounding value, given by
\begin{equation}
	\mu=\lfloor x \rfloor\, p+(\lfloor x \rfloor+\delta)(1-p).
	\label{eq:mu}
\end{equation}
Substitute \refeq{eq:mu} into \refeq{eq:variancestr1} to find
\begin{equation}
	\mathit{V}=\delta^2p(1-p).
	\label{eq:variancestr}
\end{equation}
The bias is 
\begin{equation}
	\mathit{B} = (\lfloor x \rfloor\, p+(\lfloor x \rfloor+\delta)(1-p))-x.
	\label{eq:bias}
\end{equation}
In the remainder of this section, the aforementioned expressions will be used to describe different rounding methods and their corresponding variance and bias.
\subsection{Deterministic Rounding}\label{sec:DRmodels}
For deterministic rounding methods, $p(x)=0$ or $p(x)=1$, so the variance in \eqref{eq:variancestr} is 0. These two choices are the rounding-toward-positive (ceiling) method and rounding-toward-negative (floor) method, respectively. 
Other options can be obtained by setting different $p$ for different $x$, e.g., $p(x)=1$ for $x \in [0, \tfrac{\delta}{2})$ and $p(x)=0$ for $x \in [\tfrac{\delta}{2}, \delta)$. Such rounding methods are called RN methods. RN methods vary in different tie breaking rules (different $p(\tfrac{\delta}{2})$), such as round half up, round half down, round half to even and round half to odd \cite{kahan1996ieee,8766229}. Round half up is commonly used in financial calculations \cite{cowlishaw2003decimal}. Rounding half to even is the default rounding mode used in IEEE 754 floating-point operations \cite{8766229}. It eliminates bias by rounding different numbers towards or away from zero, in such a way that the resulting biases can be compensated. In contrast, rounding half to odd is rarely employed in computations, since it will never round to zero \cite{santoro1989rounding}. The probability distribution of these methods is given in \reffig{fig:VarianceandBiasforminimizedvariance3}. 
The default RN (half to even) used in IEEE 754 floating-point operations will be further studied and compared with stochastic rounding methods in \refsec{sec:numericalstudy}. 
\subsection{CSR with Zero Bias}
\label{sec:Bias minimization}
If zero bias is required in \eqref{eq:bias}, so 
\begin{align*}
	B=\left(\lfloor x \rfloor\, p+(\lfloor x \rfloor+\delta)(1-p)\right)-x=0,
\end{align*}
we find, $p(x)=1-\tfrac{x-\lfloor x \rfloor}{\delta}$. This probability distribution is studied by \citet{nightingale1986gap,allton1989stochastic,price1993stochastic} and is widely employed in training NNs \cite{gupta2015deep,essam2017dynamic,wang2018training}. Compared to deterministic rounding methods, CSR provides an unbiased rounding result by setting a probability that is proportional to $x-\lfloor x \rfloor$. The probablity distribution, and the corresponding variance and bias are shown in \reffig{fig:VarianceandBiasforStochasticrounding}. Note that the variance is highest at the tie point.

\subsection{Improved Stochastic Rounding Method} \label{sec:Generalizedmethods}
Because RN rounds all numbers in $[-\frac{\delta}{2}, \frac{\delta}{2}]$ to 0, see \reffig{fig:VarianceandBiasforminimizedvariance3}, the gradient information in training NNs is lost in general, see \citet{hohfeld1992probabilistic} and \citet{gupta2015deep}. Though CSR is shown to provide low-precision NNs with similar accuracy as single-precision computation \cite{gupta2015deep}, gradient information is only captured partially. We propose to use $p(x)=0.5$ in \refeq{eq:strounding} for all $x\in [\lfloor x \rfloor , \lfloor x \rfloor +\delta)$. Note that $0$ is also considered during the rounding process, to minimize the chance of losing gradient information. With this choice, all the numbers are rounded up and down randomly, with equal probability. In the remainder of this paper, this rounding method is called random rounding (RR). It should be noted that the present RR is different from the one used in the Monte Carlo Arithmetic \cite{parker1997monte,parker1997monte1}, where only the inexact numbers (i.e., $x$ cannot be expressed exactly with the rounding precision) are considered in rounding. 

\reffig{fig:SR2} shows the rounding probability of RR and the corresponding rounding variance and bias. It can be seen that $\vert B\vert \leq 0.5$ is just as in RN and $V=0.25$, which is the maximum value in CSR. A summary of the main properties of the presented rounding methods can be found in \reftab{tab:summary}. It can be observed that RN provides the largest $p(x)$ for the numbers in $[-\tfrac{\delta}{2}, \tfrac{\delta}{2}]$, while RR has the smallest. 

\begin{table}[t]
	\caption{Summary of the maximum absolute bias $\vert B\vert_{\rm{max}}$, $V_{\rm{max}}$, and $p(x)$ for $x \in[-\tfrac{\delta}{2}, \tfrac{\delta}{2}]$ of RN, CSR and RR.}\label{tab:summary}
	\vskip 0.15in
	\begin{center}
		\begin{small}
			\begin{sc}
	\begin{tabular}{lcccc}
		\toprule
		Name & Figure & $\vert B \vert_{\rm{max}}$ & $V_{\rm{max}}$ & $p(x)~\text{for} ~x\in[-\tfrac{\delta}{2}, \tfrac{\delta}{2}]$ \\\midrule
		RN   &   \reffigsub{fig:VarianceandBiasforminimizedvariance3}  &      $\tfrac{\delta}{2}$                  &   $0$         &      $1$                                             \\ \rule{0pt}{2.8ex}CSR  &   \reffigsub{fig:VarianceandBiasforStochasticrounding}    &    $0$                    &  $\tfrac{\delta^2}{4}$           &     $1-\tfrac{x-\lfloor x \rfloor}{\delta}$                                              \\ \rule{0pt}{2.8ex}RR   &    \reffigsub{fig:SR2}    &    $\tfrac{\delta}{2}$                     &      $\tfrac{\delta^2}{4}$      &      $0.5$                                            \\\hline
				\end{tabular}
\end{sc}
\end{small}
\end{center}
\vskip -0.12in
\end{table}

\subsection{Rounding to a Specific Number of Fractional Bits}\label{sec:section3}
We consider fixed-point numbers for training NNs in a limited precision. In fixed-point representation, a number can be represented using the format $\mathrm{Q}[\mathrm{QI}].[\mathrm{QF}]$, where $\mathrm{QI}$ indicates the number of integer bits and $\mathrm{QF}$ denotes the number of fractional bits \cite{oberstar2007fixed}, e.g., a Q1.6 number presents a 7-bit value with 1 integer bit and 6 fractional bits. Inspired by \citet[Eq.~19]{oberstar2007fixed}, rounding a number to a specific number of fractional bits with stochastic rounding methods can be easily implemented using the operation \texttt{floor} and scaling. 
When a specific number of fractional bits is required, the rounding result can be easily achieved by multiplying with a scalar $\theta$. For instance, one fractional bit indicates a rounding precision $\delta=2^{-1}$ and the corresponding scalar is $\theta=2$. After scaling, the rounding procedure is the same as rounding-to-integer method, with $\delta=1$. The procedure for rounding to a specific number of fractional bits is given in \refag{alg:roundspdpfloat}. 

\begin{algorithm}[t]
	\caption{Round to a specific number of fractional bits.}	
	\begin{algorithmic}[1]		
		\STATE Given: Number of fractional bits: $n$. Let $\theta=2^n$. 
		\STATE For input $x$, let $\widetilde{x}= \theta x  $.
		\STATE Find 	$\mathrm{R}(\widetilde{x})$ as in \refeq{eq:strounding} with $\delta=1$.	Therefore, for CSR, $p=1-(\widetilde{x}-\lfloor\widetilde{x}\rfloor)$, while $p=0.5$ for RR.
		\STATE Scaling it back, set $\mathrm{R}(x)=\tfrac{\mathrm{R}(\widetilde{x})}{\theta}$. 
	\end{algorithmic}
	\label{alg:roundspdpfloat}
\end{algorithm}

When there are no overflow or underflow problems, we have the following property.
\begin{propt} Let $x_1, x_2, \dots, x_N \in \Real$. Then 
	\begin{align*}
		&\mathrm{R}(\mathrm{R}(\cdots \mathrm{R}(\mathrm{R}(x_1)~\mathrm{op}~x_2)~\mathrm{op}~\cdots)~\mathrm{op}~x_{N})\\&=\mathrm{R}(x_1)~\mathrm{op}~\mathrm{R}(x_2)~\mathrm{op}~\cdots~\mathrm{op}~\mathrm{R}(x_{N}),
	\end{align*}
	where $N$ indicates the number of terms for $\mathrm{op}\in\{+,-\}$.
	\label{prop: roundsum}
\end{propt}
\begin{proof}
	Let $p_1=p(x_1)$, $p_2=p(x_2)$ in \refeq{eq:strounding}. Then
	\begin{align*}
		&\mathrm{R}(x_1)+\mathrm{R}(x_2)\\&=\begin{cases}
			\lfloor x_1 \rfloor +\lfloor x_2 \rfloor,          \\ \quad\text{with probability~} p_1p_2, \\
			\lfloor x_1 \rfloor +\lfloor x_2 \rfloor +\delta, \\  \quad\text{with probability~} (1-p_1)p_2 +p_1(1-p_2),\\
			\lfloor x_1 \rfloor +\lfloor x_2 \rfloor +2\delta, \\ \quad\text{with probability~} (1-p_1)(1-p_2).
		\end{cases}
	\end{align*}	
	We also have
	\begin{align*}
		&\mathrm{R}\big(\mathrm{R}(x_1)+ x_2\big)\\
		&=\begin{cases}
			\lfloor \lfloor x_1 \rfloor+ x_2 \rfloor =  \lfloor x_1 \rfloor+\lfloor x_2 \rfloor, \\ \quad\text{with probability}~p_1p_2,\\
			\lfloor \lfloor x_1 \rfloor+ \delta+ x_2\rfloor = \lfloor x_1 \rfloor+\lfloor x_2 \rfloor +\delta, \\\quad\text{with probability}~(1-p_1)p_2 +p_1(1-p_2),\\
			\lfloor \lfloor x_1 \rfloor+ \delta+ x_2\rfloor+\delta=  \lfloor x_1 \rfloor+\lfloor x_2 \rfloor +2\delta, \\ \quad\text{with probability}~(1-p_1)(1-p_2),
		\end{cases} \\                                                                                                                           
		&=\mathrm{R}(x_1)+ \mathrm{R}(x_2).
	\end{align*}	
	For summation of $N$ terms,
	\begin{align*}
		&\mathrm{R}(\mathrm{R}(x_1)+x_2)+\cdots)+x_{N})\\
		&~~~~~=\begin{cases}
			\lfloor x_1 \rfloor + \lfloor x_2 \rfloor +\cdots +\lfloor x_{N} \rfloor, \\\quad\text{with probability}~p_1p_2\cdots p_{N},\\
			\lfloor x_1 \rfloor + \lfloor x_2 \rfloor + \cdots +\lfloor x_{N} \rfloor +\delta, \\
			\quad\text{with probability}~(1-p_1)p_2\cdots p_{N}\\+p_1(1-p_2)\cdots p_{N}+\cdots+p_1p_2\cdots (1-p_{N}),\\
			\quad \quad \quad \quad \vdots \\
			\lfloor x_1 \rfloor + \lfloor x_2 \rfloor + \cdots +\lfloor x_{N} \rfloor +N\delta,\\
			\quad\text{with probability}~(1-p_1)(1-p_2)\cdots (1-p_{N}),\\
		\end{cases}\\
		&~~~~~=\mathrm{R}(x_1)+ \mathrm{R}(x_2)+\cdots+\mathrm{R}(x_{N}).
	\end{align*}
	The above relation also holds for subtraction.
\end{proof}

\section{Mathematical Operations in Training NNs}\label{sec:section4}
\begin{table*}[t]
	\caption{Sum of the absolute bias $\vert B\vert$ and number of zeros $N_z$ of RN, CSR and RR for computing dot products, over the total number of dot product operations ($N_{\rm{max}}$), where $N$ is the number of elements in $\boldsymbol{x}$ and $\boldsymbol{y}$. Largest values of $\vert B\vert$ and smallest values of $N_z$ are marked in bold. }\label{tab:dotproduct}
	\vskip 0.15in
	\begin{center}
		\begin{small}
			\begin{sc}
				\begin{tabular}{llrrrrrrrr}
					\toprule
					\multicolumn{2}{l}{\rule{0pt}{2.1ex}$N$}          & \multicolumn{4}{c}{100}    & \multicolumn{4}{c}{200}   \\ \midrule
					\multicolumn{2}{l}{\rule{0pt}{2.1ex}$N_{\rm{max}}$} & $1000$           & $2000$           & $3000$           & $4000$           & $1000$          & $2000$           & $3000$  & $4000$ \\\hline
					\multirow{3}{*}{$\vert B \vert$}   & \rule{0pt}{2.8ex}RN       & \phantom{1}$5.0$           & $10.2$          & $15.2$          & $20.6$          & \phantom{1}$3.6$           & \phantom{1}$7.2$           & $10.9$          & $14.6$          \\
					& CSR       & \phantom{1}$7.4$           & $14.7$          & $22.2$          & $29.6$          & \phantom{1}$5.3$           & $10.6$          & $15.7$          & $21.2$          \\
					& RR & $\boldsymbol{12.2}$          & $\boldsymbol{24.0}$          & $\boldsymbol{35.7}$           & $\boldsymbol{47.9}$          & \phantom{1}$\boldsymbol{9.4}$           & $\boldsymbol{18.5}$          & $\boldsymbol{27.6}$          & $\boldsymbol{36.8}$     \\\hline
					\multirow{3}{*}{$N_{z}$}           & \rule{0pt}{2.4ex}RN       & $1000$           & $2000$           & $3000$           & $4000$           & $1000$           & $2000$           & $3000$           & $4000$           \\
					& CSR       & \phantom{1}$132$            & \phantom{1}$268$            & \phantom{1}$393$            & \phantom{1}$520$            & \phantom{1}$198$            & \phantom{1}$400$            & \phantom{1}$592$            & \phantom{1}$768$            \\
					& RR      & \phantom{12}$\boldsymbol{50}$             & \phantom{12}$\boldsymbol{85}$             & \phantom{1}$\boldsymbol{131}$            & \phantom{1}$\boldsymbol{182}$            & \phantom{12}$\boldsymbol{64}$             & \phantom{1}$\boldsymbol{121}$            & \phantom{1}$\boldsymbol{187}$            & \phantom{1}$\boldsymbol{244}$            \\\hline
				\end{tabular}
			\end{sc}
		\end{small}
	\end{center}
	\vskip -0.12in
\end{table*}

In this section, we will briefly explore the main mathematical operations when training NNs. For details see \citet{higham2019deep}. The training procedure is generally divided into three parts: forward propagation, backward propagation and weights updating. For $m$ training datasets, in the forward propagation, the intermediate variable and the output of the $i$th layer, $Z^{[i]}$ and $A^{[i]}\in \Real^{n_i\times m}$, respectively, are computed using 
\begin{equation}\label{eq:forwardprop}
	Z^{[i]}=W^{[i]}A^{[i-1]}+b^{[i]},\quad
	A^{[i]}=\sigma(Z^{[i]}),
\end{equation}
where $n_i$ denotes the number of neurons for the $i$th layer, $\sigma$ indicates the activation functions, $W^{[i]}\in R^{n_i\times n_{i-1}}$ and $b^{[i]}\in R^{n_i\times m}$ denote the weights and biases matrix, respectively. In the backward propagation, the parameters' gradients are computed with respect to the cost function, that is $\frac{\partial J}{\partial W}$ and $\frac{\partial J}{\partial b}$. Here, a binary cross-entropy loss function is considered as the cost function, given by 
\begin{align*}
	J(A^{[l]},Y)&=\tfrac{1}{m}\sum_{j=1}^{m}L_j(a_j,y_j),\\
	L_j(a_j,y_j)&=-y_j\log(a_j)-(1-y_j)\log(1-a_j),
\end{align*}
where $y_j$ is the observed value of the $j$th training dataset and $a_j$ refers to the output of an $l$-layer NN for the $j$th dataset in $A^{[l]}$. For a two-layer NN, we have
\begin{align*}
	\frac{\partial J}{\partial W^{[2]}} &=\frac{\partial J}{\partial A^{[2]}}\frac{\partial A^{[2]}}{\partial Z^{[2]}}\frac{\partial Z^{[2]}}{\partial W^{[2]}},\\
	\frac{\partial J}{\partial {b}^{[2]}} &=\frac{\partial J}{\partial A^{[2]}}\frac{\partial A^{[2]}}{\partial Z^{[2]}}\frac{\partial Z^{[2]}}{\partial{b}^{[2]}}.
\end{align*}
If the sigmoid activation function is applied in the second layer, these two derivatives can be simplified as
\begin{subequations}\label{eq:backprop}
	\begin{align}
		\frac{\partial J}{\partial W^{[2]}} &=\tfrac{1}{m}(A^{[2]}-Y)A^{[1]}, \label{eq:backprop1}\\
		\frac{\partial J}{\partial {b}^{[2]}} &=\tfrac{1}{m}\sum_{j=1}^{m} (A^{[2]}-Y)\label{eq:backprop2}.
	\end{align}
\end{subequations}
Further, $\frac{\partial J}{\partial W^{[1]}}$ and $\frac{\partial J}{\partial {b}^{[1]}}$ can be calculated accordingly. Finally, the parameters for the $k$th iteration in the gradient descent method are updated as follows:
\begin{equation}\label{eq:updates}
	W_{k}^{[i]} =W_{k-1}^{[i]}-\alpha \frac{\partial J}{\partial W_{k-1}^{[i]}},\quad
	{b}_{k}^{[i]} ={b}_{k-1}^{[i]}-\alpha \frac{\partial J}{\partial {b}_{k-1}^{[i]}},
\end{equation}
where $\alpha$ indicates the learning rate, that can be considered as a hyperparameter. Overall, the main mathematical operations in training NNs are matrix multiplication, summation and subtraction. 

\section{Numerical Experiments}
\label{sec:numericalstudy}
In this section, all the rounding modes are studied and compared on dot product operation. Next, we train NNs using limited precision with different rounding modes. It should be noted that all the experiments are done with fixed-point numbers using Matlab, but the method is general and can be repeated for other NNs and software. Because the fixed-point computation and rounding are done in software, the computation speed is hard to estimate, only the accuracy and convergence rate will be studied. 

\subsection{Dot Product Computation with Limited Precision}\label{sec:dotproduct}
Similar as \citet[Section 3.2]{gupta2015deep}, for the multiply and accumulate operation, rounding is applied after the accumulation of all the sums. To avoid overflow problems, rounding in \refeq{eq:backprop1} is applied after the division over the total number of datasets. In matrix multiplication, each entry in the resulting matrix is the dot product of a row in the first matrix and a column in the second matrix. For two vectors $\boldsymbol{x}, \boldsymbol{y} \in \Real^N$, we have
\begin{align}
	&\mathrm{R}\big(\frac{\mathrm{R}(\boldsymbol{x})\cdot\mathrm{R}(\boldsymbol{y})}{N}\big)\\&=\mathrm{R}\big(\frac{\mathrm{R}(x_1)\mathrm{R}(y_1)+\mathrm{R}(x_2)\mathrm{R}(y_2)+\nonumber \dots +\mathrm{R}(x_{N})\mathrm{R}(y_{N})}{N}\big),
	\label{eq:dotproduct_round}
\end{align}
where $N$ is the number of elements in $\boldsymbol{x}$ and $\boldsymbol{y}$. To simulate the influence of zeros of matrix multiplication in training NNs, $\boldsymbol{x}$ and $\boldsymbol{y}$ are generated using a random number generator, where $\boldsymbol{x}$ and $\boldsymbol{y}$ are uniformly distributed in $[-\tfrac{\delta}{2},\tfrac{\delta}{2}]$ and $[0,10]$, respectively. All the numbers are represented using fixed-point representation, with 16-bit word (16W) length, 8-bit fractional (8F) length and 1-bit for the sign. 

\reftab{tab:dotproduct} shows the sum of absolute bias $\vert B \vert$ and the total number of zeros ($N_z$) over the total number of dot product operations ($N_{\rm{max}}$), where the largest value of $\vert B\vert$ and smallest $N_z$ are marked in bold. It can be observed that the smallest bias and largest number of zeros are always obtained by RN, while the largest bias and smallest number of zeros are always achieved by RR. CSR is in between these two methods. Based on these observations, RR may lead to the fastest convergence speed, while RN may result in the slowest convergence speed in training NNs. 

\subsection{Training NNs using Limited Precision}
In this section, a two-layer NN is trained with limited precision using the MNIST database. The MNIST is a large database of $10$ handwritten digits (from $0$ to $9$), containing $60,000$ training images and $10,000$ test images. In the numerical experiments, a binary classification problem is considered. The NNs will be trained on two sets of data. The first set of data is comprised of digits 6 and 9, resulting in $11,867$ training images and $1,967$ testing images. The second one contains digits 3 and 8, resulting in $11,982$ training images and $1,984$ testing images. Similarly as in the paper from \citet{gupta2015deep}, the pixel values are normalized to $[0,1]$. A two-layer NN is built with ReLu activation function in the hidden layer and sigmoid activation function in the output layer. The hidden layer contains 100 units. In the backward propagation, a binary cross-entropy loss function is optimized using the batch gradient descent method. The weights matrix ($W$) is initialized based on Xavier initialization \cite{glorot2010understanding} and the bias ($b$) is initialized as a zero matrix. The learning rate is $\alpha=0.1$ in \eqref{eq:updates}. To make the experiments repeatable, the random numbers are generated using the same seed for each rounding mode. Additionally, the default decision threshold is set for interpreting probabilities to class labels, that is 0.5, since the sample class sizes are almost equal \cite{chen2006decision}. Specifically, class 1 is defined for those predicted scores larger than or equal to $0.5$.
\begin{figure*}[t]
	\vskip 0.15in
	\centering
	\subfloat{\label{fig:etrain69_16w10f_v1}\includegraphics[width=0.41\textwidth]{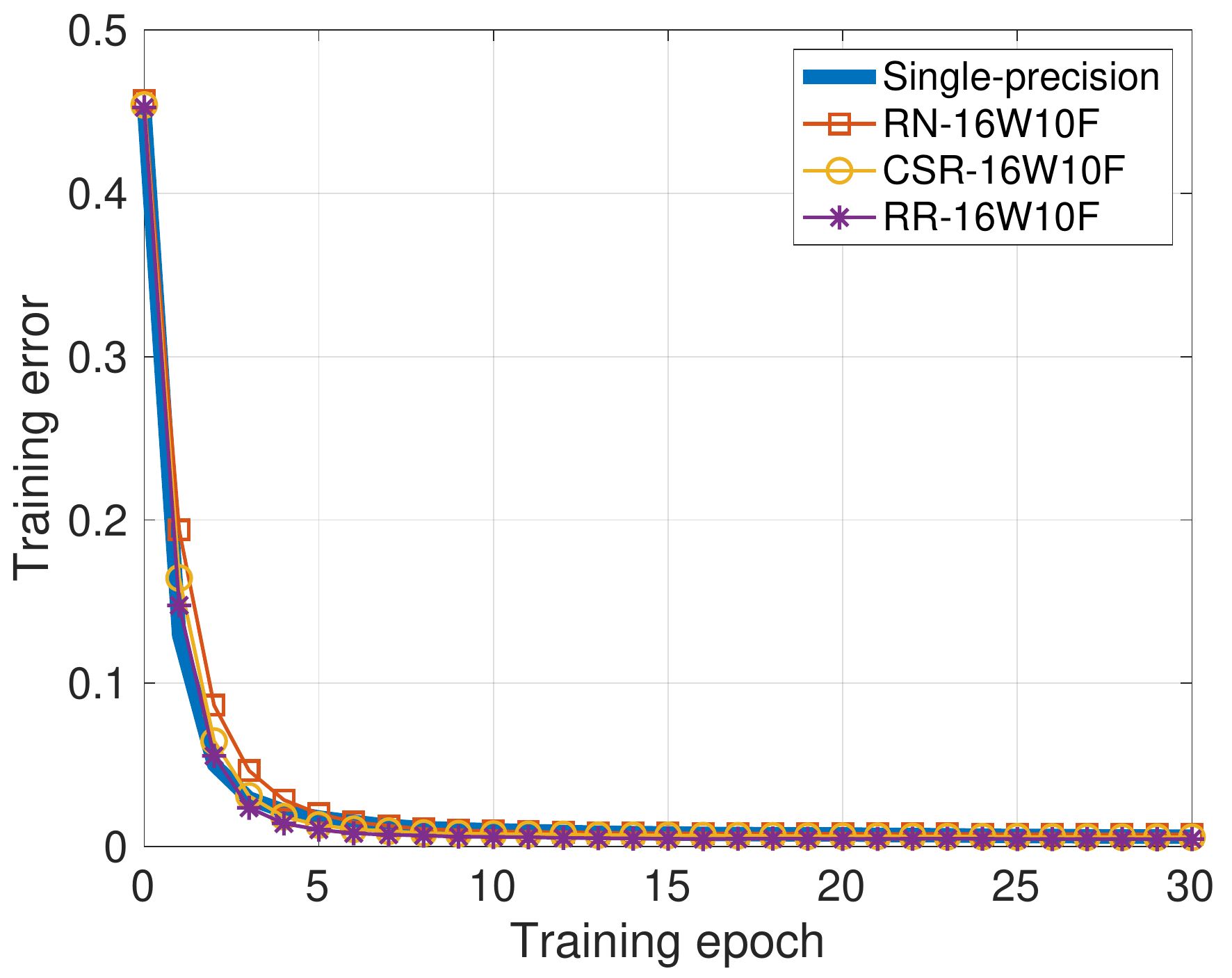}}	\subfloat{\label{fig:epre69_16w10f_Nz}\includegraphics[width=0.41\textwidth]{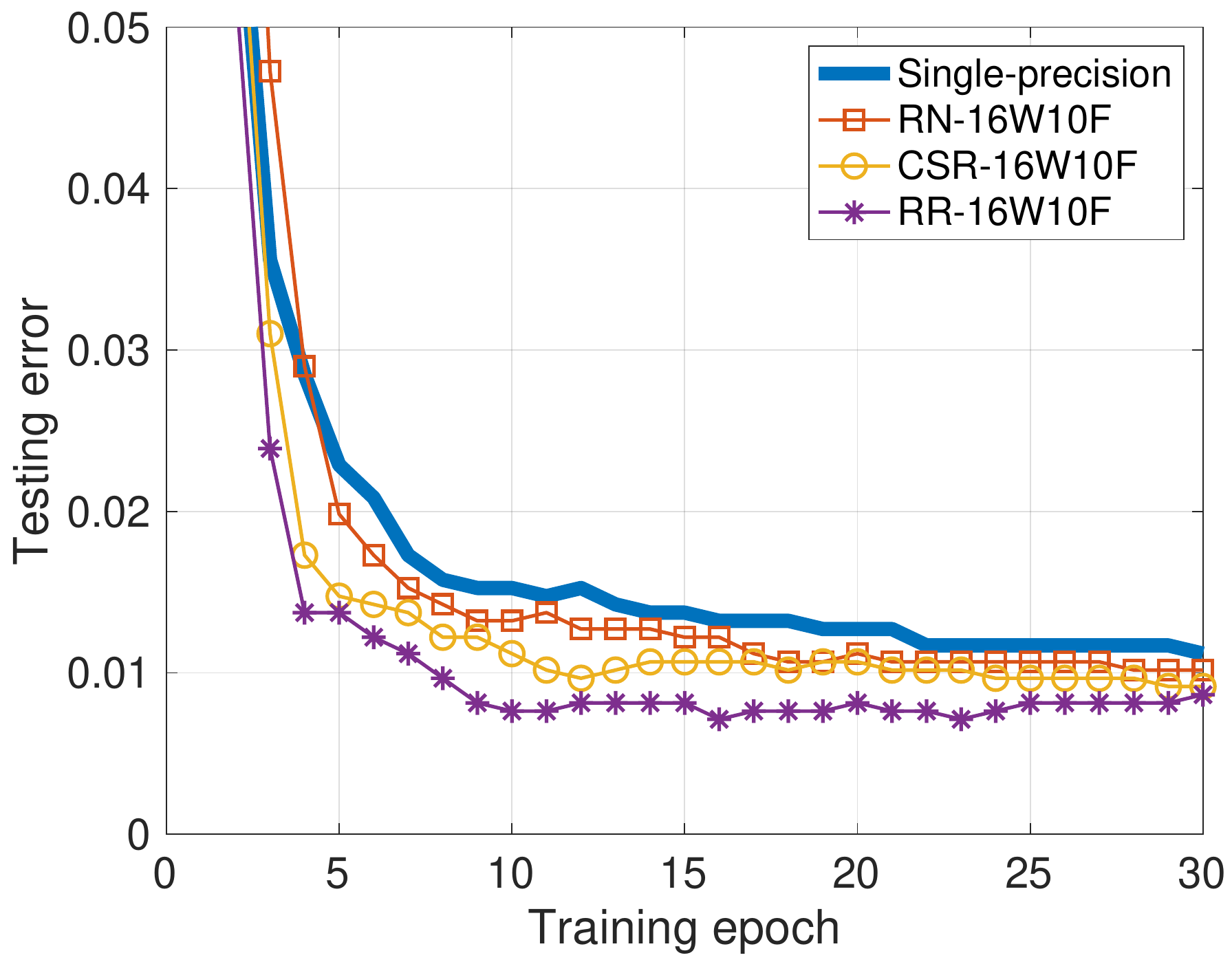}}
	\caption{Digits 6 and 9: Training error (left) and testing error (right) of the NN trained using fixed-point numbers with 16W and 10F using RN, CSR and RR; the blue solid baseline is obtained by single-precision computation in both figures.}
	\label{fig:69_16w10f}
	\vskip -0.12in
\end{figure*} 
\begin{figure*}[t]
	\vskip 0.15in
	\centering
	\subfloat{\label{fig:etrain69_16w8f_scaled}\includegraphics[width=0.41\textwidth]{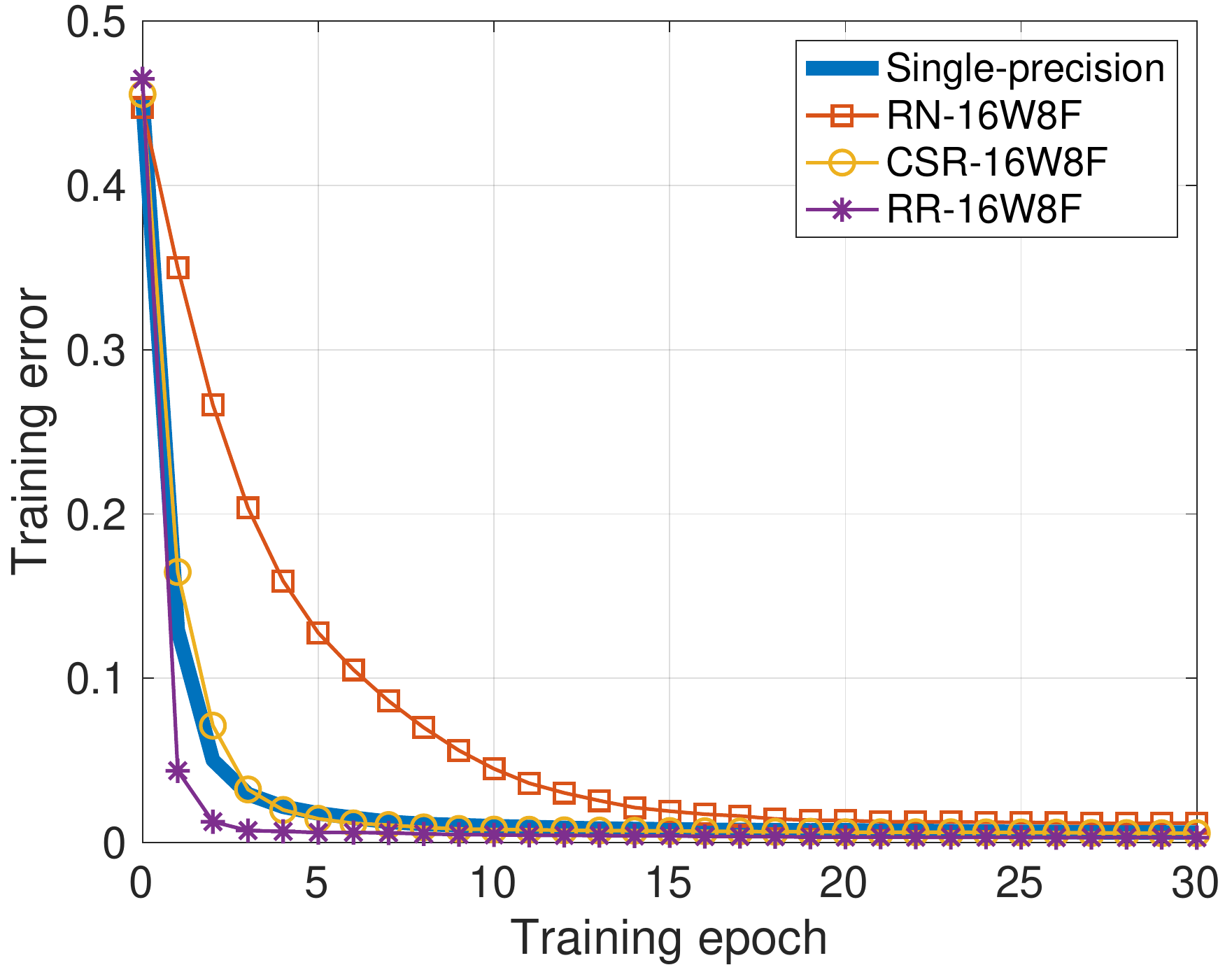}}
	\subfloat{\label{fig:epre69_16w8f_scaled_zoomed}\includegraphics[width=0.41\textwidth]{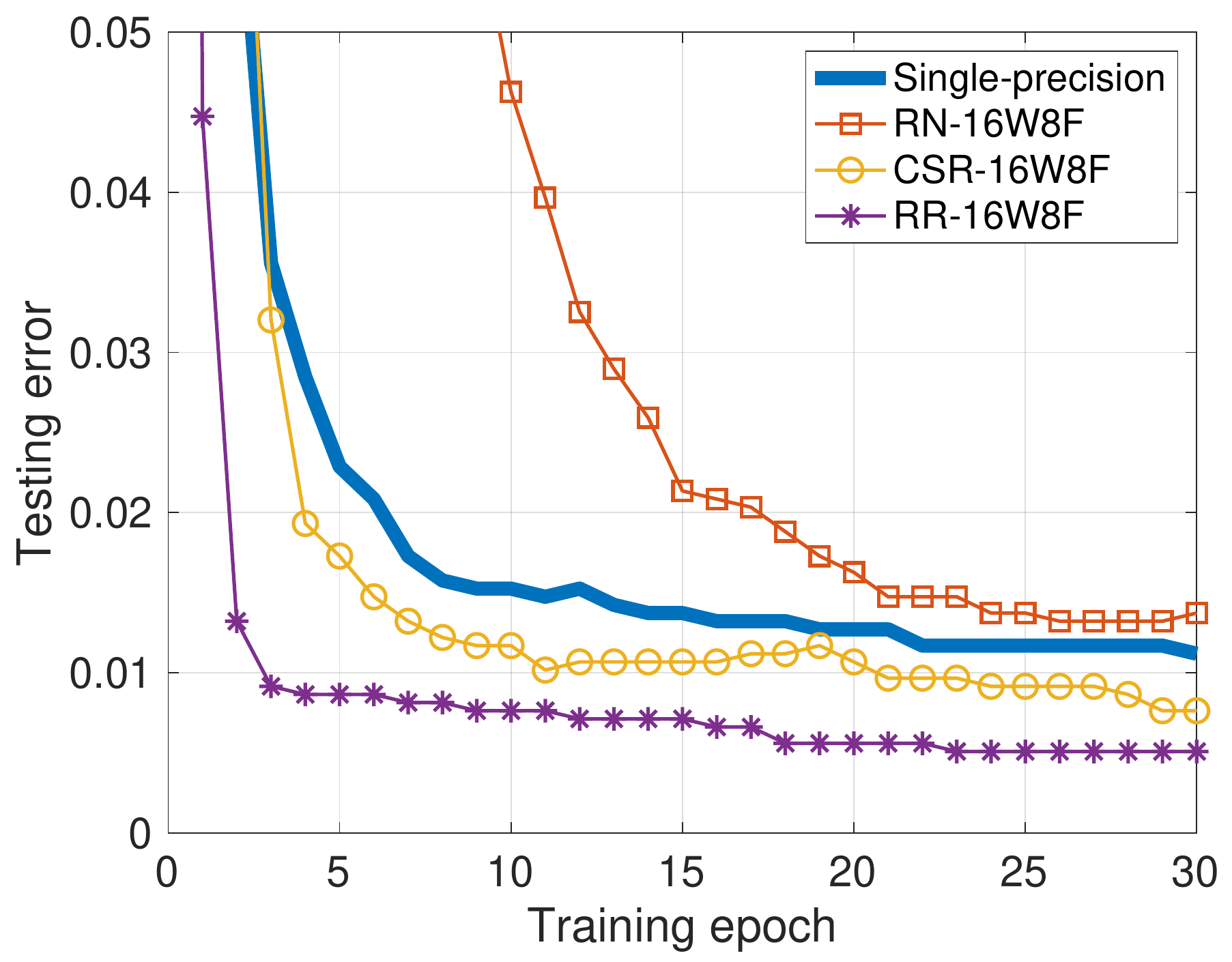}}
	\caption{Digits 6 and 9: Training error (left) and testing error (right) of the NN trained using fixed-point numbers with 16W and 8F using RN, CSR and RR; the blue solid baseline is obtained by single-precision computation in both figures.}
	\label{fig:NN69_16w8f_normalized}
	\vskip -0.12in
\end{figure*}
\begin{figure*}[t]
	\vskip 0.15in
	\centering
	\subfloat{\includegraphics[width=0.41\textwidth]{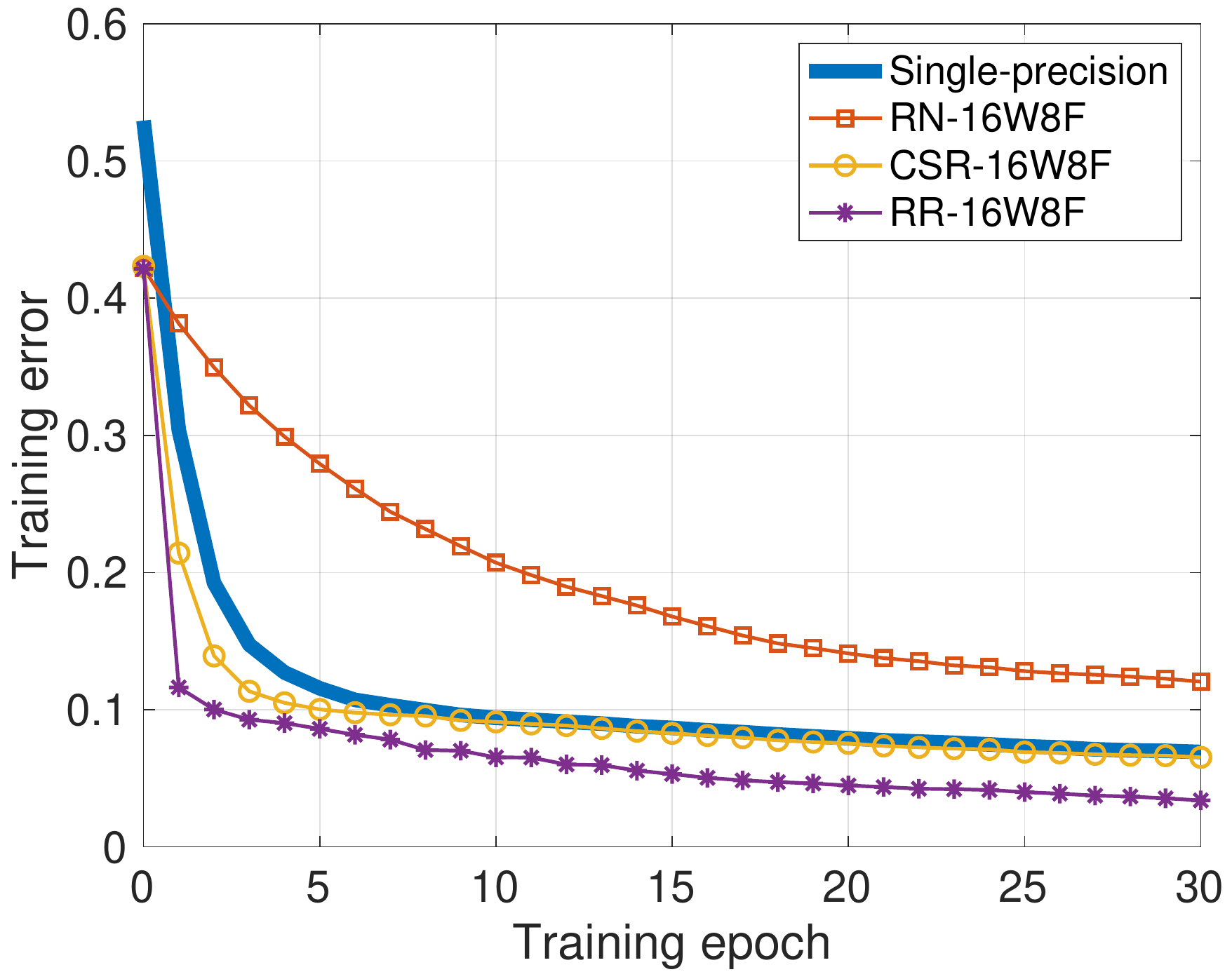}}
	\subfloat{\includegraphics[width=0.41\textwidth]{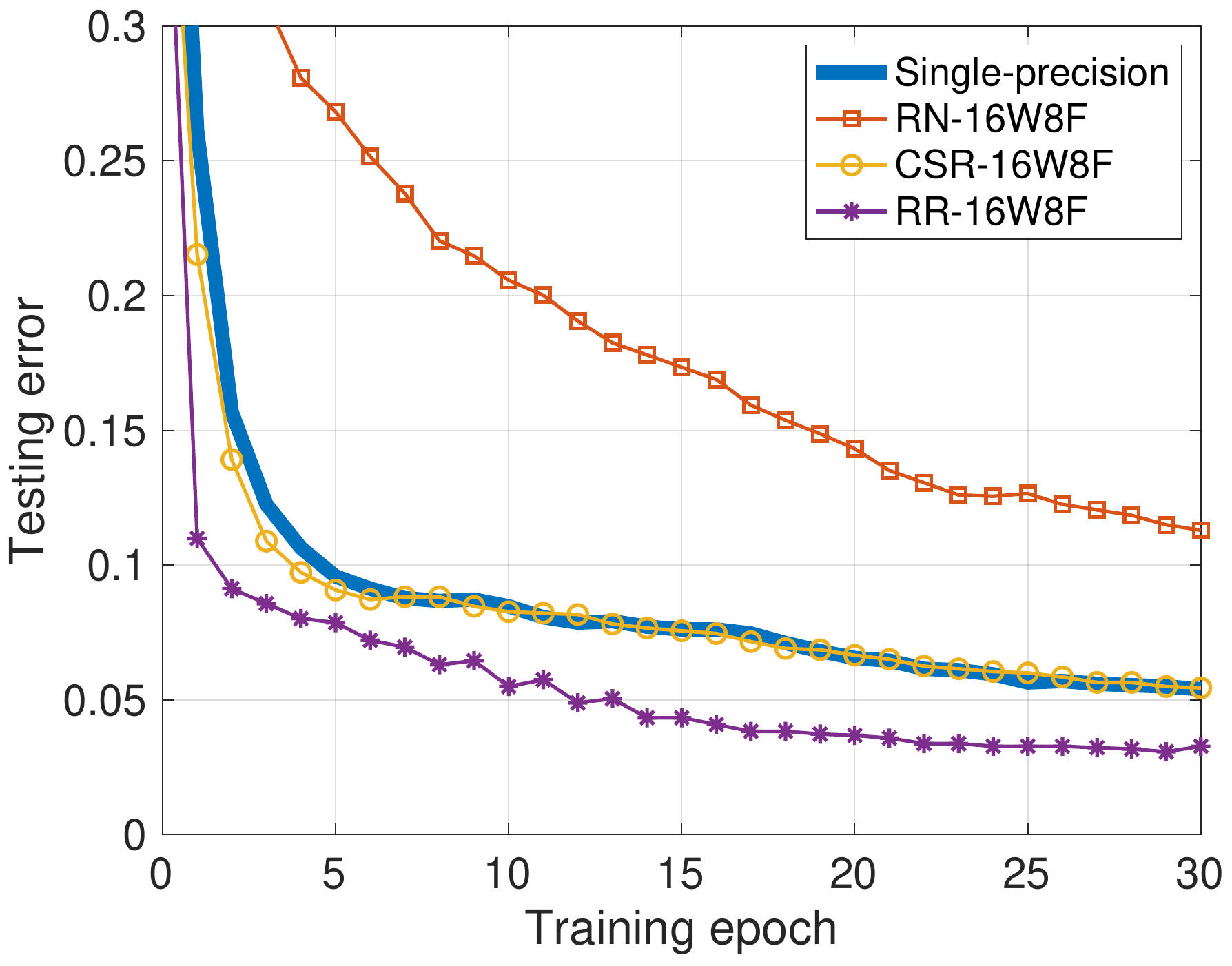}}
	\caption{Digits 3 and 8: Training error (left) and testing error (right) of the NNs trained using fixed-point numbers with 16W and 8F using RN, CSR and RR; the blue solid baseline is obtained by single-precision computation in both figures.}
	\label{fig:NN38_16w8f_normalized}
	\vskip -0.12in
\end{figure*}
\begin{figure*}[h!]
	\vskip 0.15in
	\centering
	\subfloat[Single-precision]{\label{fig:outsingle_38}\includegraphics[width=0.41\textwidth]{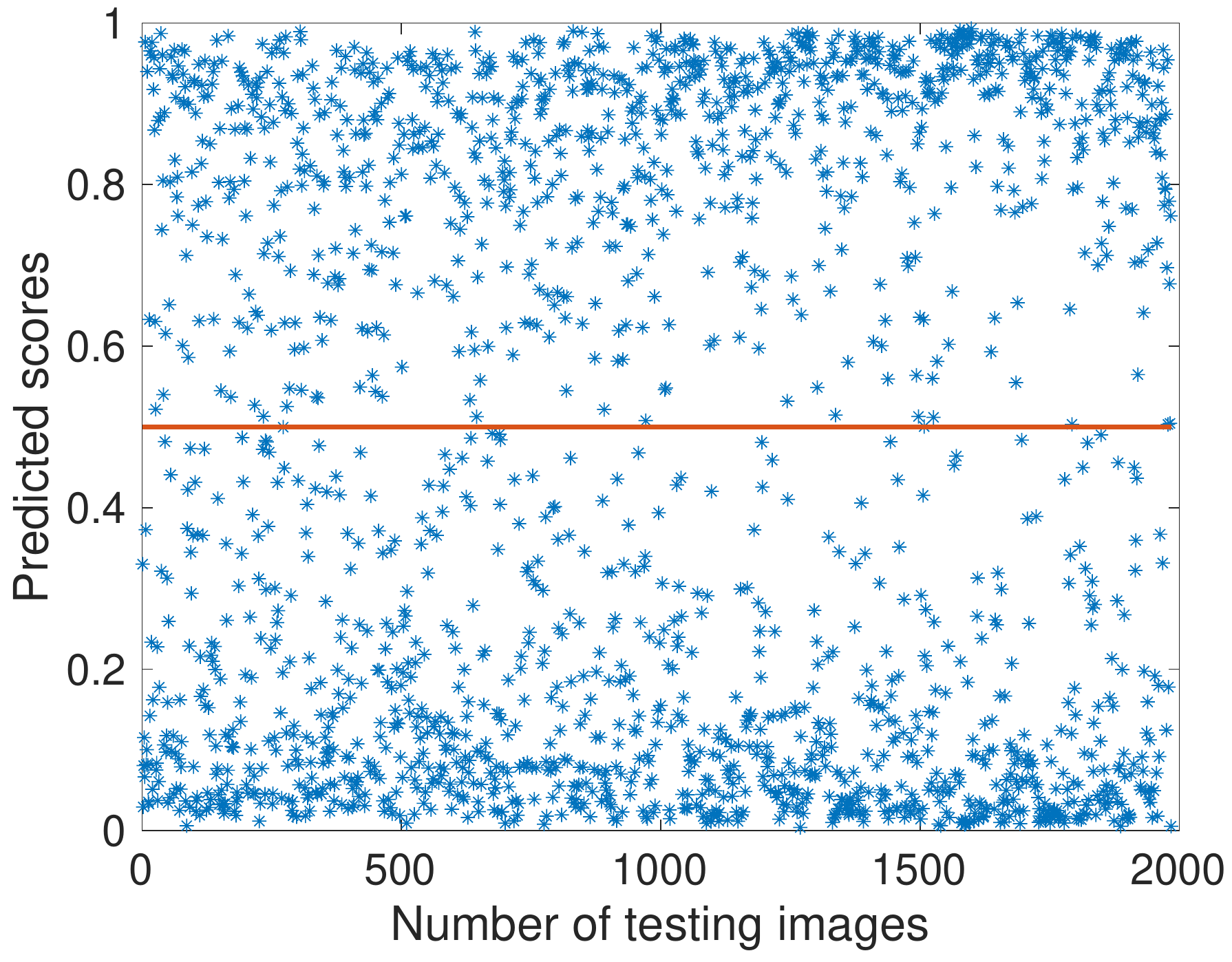}}
	\subfloat[RN (16W8F)]{\label{fig:outCR_38}\includegraphics[width=0.41\textwidth]{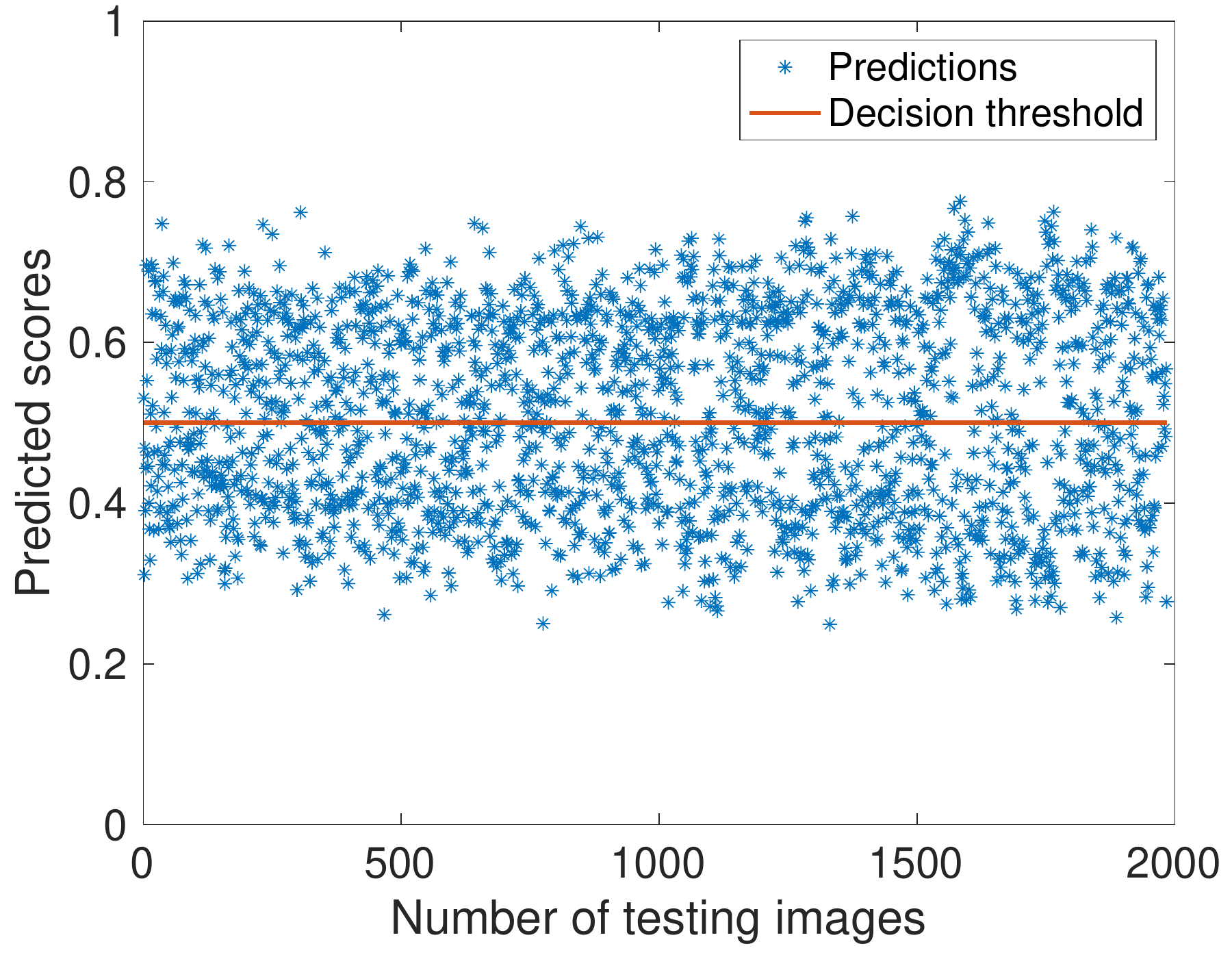}}\\
	\subfloat[CSR (16W8F)]{\label{fig:eoutCSR_38}\includegraphics[width=0.41\textwidth]{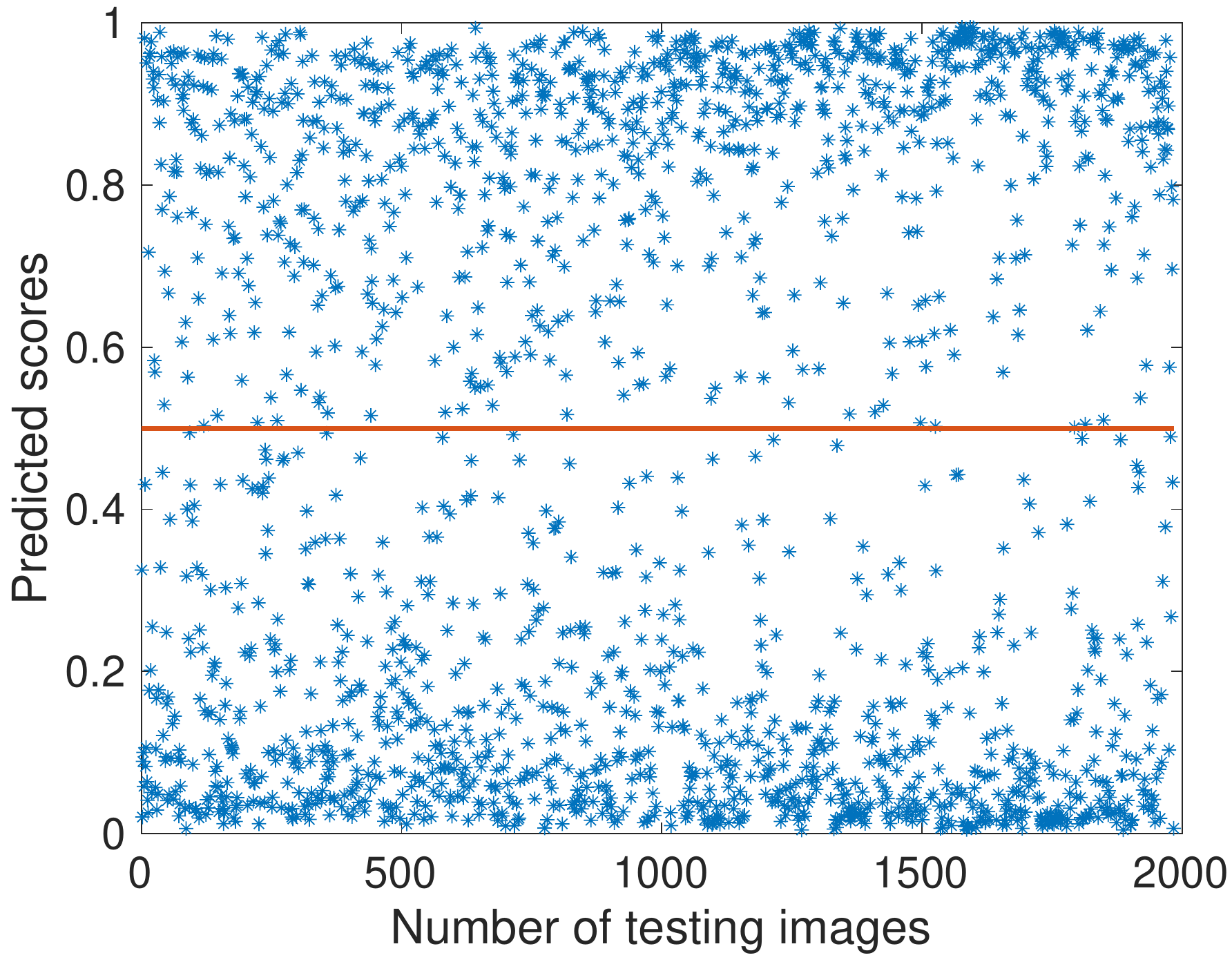}}
	\subfloat[RR (16W8F)]{\label{fig:outRR_38}\includegraphics[width=0.41\textwidth]{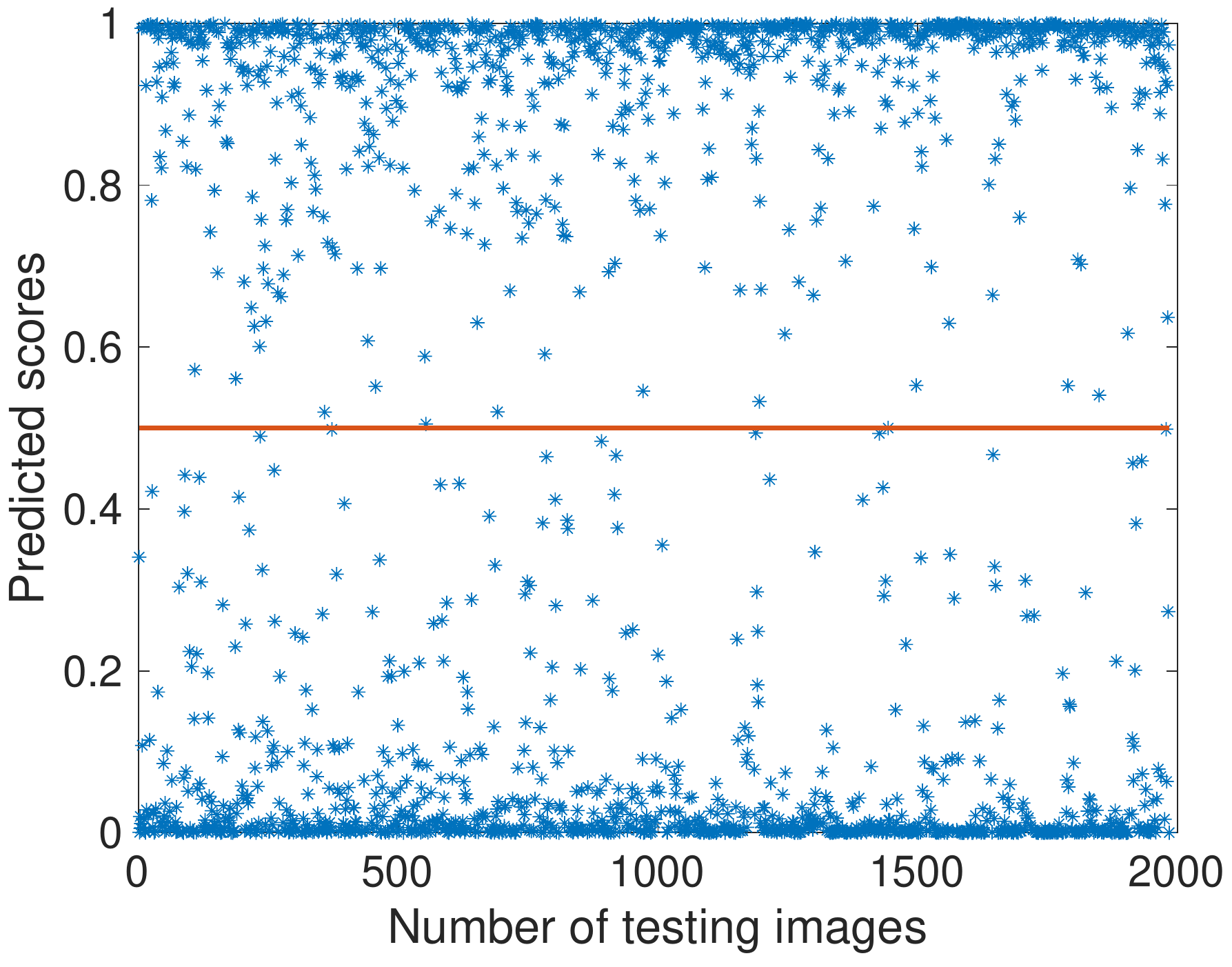}}
	\caption{Digits 3 and 8: Predicting outputs of the NNs trained using single-precision computation (a), 16W8F fixed-point numbers that are rounded using RN (b), CSR (c) and RR (d), after 30 training epochs.}
	\label{fig:predictingoutput_38}
	\vskip -0.12in
\end{figure*}

In the forward propagation, \refeq{eq:forwardprop}, the rounding process is considered for each operation, e.g., $\mathrm{R}(\mathrm{R}(W^{[i]}A^{[i-1]})+\mathrm{R}({b}^{[i]}))$. In the backward propagation, e.g., \refeq{eq:backprop}, the rounding process is applied after the accumulation of all the sums, e.g., based on \refprop{prop: roundsum}, the rounded value of \eqref{eq:backprop2} is $\mathrm{R}(\tfrac{1}{N}\sum_{i=1}^{N}\mathrm{R}(A^{[2]}-Y))$, for which the rounding process after the sum is not considered to avoid the overflow problem. The finite sum operation is computed using the fixed-point toolbox in Matlab and the number of integer bits of growth after the sum operation is $\lceil \log_2(N)\rceil$. As a result, the total word length after the sum operation will be approximately 30 when $N=11982$. It should be noted that this number can be reduced by employing the mini-batch gradient descent method, in such a way that the batch size can be adjusted according to the hardware requirements, e.g., when the mini-batch size is 100, only 23 bits are required. Additionally, a small mini-batch size will also help avoid overflow problem for rounding after the finite sum. For the rest of the calculations in the backward propagation, values are rounded after each operation. Further, the baseline is obtained by evaluating the NN using single-precision (floating-point) computation. It should be noted that all the hyperparameters are kept the same for the numerical experiments in this section. 

In this section, the results of the NNs trained using RR are compared to the NNs trained using RN and CSR. The NNs are trained using fixed-point numbers with 16-bit word length (16W) and different fractional length. First, the NNs are trained using the digits 6 and 9. \reffig{fig:69_16w10f} shows the training errors and predicting errors obtained by the fixed-point numbers with 16W and 10-bit fractional length (10F). It can be observed that the fixed-point computations preserve similar training and predicting accuracy as single-precision computation for all the rounding modes in \reffig{fig:69_16w10f}. The single-precision float baseline obtains a test error of $1.12\%$ after 30 training epochs. The test errors obtained by RN and CSR are $0.92\%$, resulting in a slightly higher classification accuracy than single-precision computation, while the test error achieved by RR is $0.86\%$. In \reffig{fig:69_16w10f}, it can be seen that the convergence speed is decreasing with increasing number of zeros in the matrix multiplications during the parameter updating procedure. As it is expected in \refsec{sec:dotproduct}, RN leads to the slowest convergence speed among all the rounding modes, while RR is fastest. Comparing \reftab{tab:dotproduct} and \reffig{fig:69_16w10f}, it can be seen that the convergence speed increases when $N_z$ decreases. 

Similar simulations are done with the fractional length reduced to 8 bits, while the word length is kept at 16 bits. \reffig{fig:NN69_16w8f_normalized} shows the training error and testing error of the NNs. Again, the training and predicting accuracy of single-precision computation are preserved by all the rounding modes. However, the convergence rate of RN is strongly decreased compared to single precision. Again, RR is fastest. Further, the NN trained using RR provides a testing error of $0.76\%$ after 9 training epochs, while the NN trained using CSR achieves the same testing error after 30 training epochs. Overall, RR achieves the NN with fastest convergence rate and higher training accuracy than that obtained by single precision. 

To investigate the validity of our observations, we have repeated the training process on the MNIST dataset with labels 3 and 8, while all other parameters are kept the same. \reffig{fig:NN38_16w8f_normalized} shows training and testing errors of the NNs trained using fixed-point numbers with 16W and 8F. The baseline of testing error obtained by single-precision computation is $5.44\%$, represented by the blue solid line. From \reffig{fig:NN38_16w8f_normalized}, a large degradation from the single-precision baseline can be observed in both convergence rate and classification accuracy, when RN is applied. As stated by \citet{hohfeld1992probabilistic}, \citet{gupta2015deep}, and shown in \reftab{tab:dotproduct}, with RN, most parameter updates are rounded to zero. CSR provides an NN with almost the same testing error as the single-precision computation. Again, RR achieves the most accurate NN after 30 training epochs, with a testing error of $3.28\%$. It can be observed that RR already provides an NN with a testing error of $4.89\%$ after 12 training epochs. 

\reffig{fig:predictingoutput_38} shows a comparison of the predicted outputs of the NNs trained using single-precision computation (\reffig{fig:outsingle_38}), 16W8F fixed-point numbers with RN (\reffig{fig:outCR_38}), CSR (\reffig{fig:eoutCSR_38}) and RR (\reffig{fig:outRR_38}), after 30 training epochs. The predicted outputs of the NN trained using CSR are very similar to those obtained by single-precision computation (as shown in \reffig{fig:NN38_16w8f_normalized}). The predicted outputs of the NN trained using RN suffers from serious data distortion. However, it can be seen from \reffig{fig:outRR_38} that the NN trained using RR provides even better information than single-precision computation. 

It turns out that zero rounding bias may not be necessarily required in training NNs. A large probability of rounding numbers to zero generally leads to larger degradation in classification accuracy and slower convergence rate. It may also cause the output of NNs to be biased, see, e.g., \reffig{fig:predictingoutput_38}. RN provides NNs with similar accuracy as CSR and RR, when the rounding precision is small enough, see. e.g., \reffig{fig:69_16w10f}. Furthermore, similarly as in the paper of \citet{gupta2015deep}, a further reduction in precision will make all the rounding modes fail to capture the gradient information during the parameter updating procedure. Generally, RR achieves the NNs with the highest classification accuracy and fastest convergence rate (at least twice faster than CSR). Moreover, RR may be a better option than CSR in hardware implementation, since it uses a constant rounding probability, compared to the linear relation in CSR. 

\section{Conclusion}
\label{sec:conclusion}
Rounding is an essential step in many computations; round-off errors are unavoidable. In the training of neural networks (NNs) with limited precision, deterministic rounding methods generally suffer from the loss of gradient information, while conventional stochastic rounding (CSR) captures part of the gradients. In this paper, we propose an improved stochastic rounding method named Random Rounding (RR). RR applies a constant rounding probability, that may simplify the hardware implementation compared to CSR. 

Numerical experiments have been performed in dot product operations and show that fewer numbers are rounded to zeros in RR. Further, low-precision fixed-point computations have been applied to train NNs with different rounding modes for binary classification problems. It has been shown that RR achieves NNs with 16-bit fixed-point representations that can provide higher classification accuracy and faster convergence rate than those using rounding-to-the-nearest, CSR and single-precision floating-point computation. Due to the fixed probability to round numbers in RR, the computational complexity is low. 

The potential of RR in other areas is still to be studied. Further study may focus on the training of conventional NNs for multi-class classification problems and exploring the maximum allowable rounding bias in training NNs. The influence of RR on different optimization methods is also to be studied.

\section*{Acknowledgments}
This research was funded by the EU ECSEL Joint Undertaking under grant agreement no.826452 (project Arrowhead Tools).

\end{document}